\documentclass[runningheads]{llncs}

\usepackage[T1]{fontenc}

\usepackage{graphicx}
\usepackage{mathastext}
\usepackage{amsmath}
\usepackage{array}
\usepackage{color,soul}
\soulregister\ref7
\soulregister\cite7

  \usepackage[caption=false,font=footnotesize]{subfig}

\begin{document}

\title{VSTaI: Design and Characterization of Variable-Stiffness Tactile Interfaces Based on 3D-Printed Structured Fabrics}

\author{Yiting Mo\inst{1} \and Xinyuan Mao\inst{2} \and Jashan Preet Singh\inst{3}  \and Fernando Bello\inst{1,4}} 

\titlerunning{VSTaI: Variable-Stiffness Tactile Interfaces}

\institute{Duke-NUS Medical School, National University of Singapore, Singapore 
\email{yt.mo@duke-nus.edu.sg, f.bello@duke-nus.edu.sg}\\
\and College of Design and Engineering, National University of Singapore, Singapore \\
\and Mechanobiology Institute, National University of Singapore, Singapore \\
\and
Department of Surgery and Cancer, Imperial
College, London, UK}

\maketitle              

\begin{abstract}

Realistic palpation training requires reliable rendering of soft tissue stiffness changes in real time, which is difficult to achieve with conventional simulators. 
This paper presents a compact, variable-stiffness tactile interface (VSTaI) based on vacuum-induced jamming of 3D-printed structured fabrics. 
A vacuum-sealed fabric layer is sandwiched between two silicone layers, and stiffness is tuned by regulating internal pressure.
Four fabric patterns with different geometric parameters were fabricated and evaluated using force–indentation tests under atmospheric and vacuum conditions. 
Across the tested pattern and geometry combinations, vacuum jamming increased stiffness significantly, producing an effective modulus from sub-megapascal to megapascal levels.
Specifically, one configuration exhibited a stiffness increase of up to $140\%$ under the jammed state.
Circular chainmail patterns provided the most spatially uniform distribution of tactile stiffness, while denser geometries reached higher peak stiffness. VSTaI was also shown to exhibit excellent conformability to the underlying geometry.
These results support structured-fabric jamming as a practical approach for shape-conformable, tunable-stiffness displays aimed at physical examination training.

\end{abstract}

\section{Introduction}

Physical examination, particularly palpation, remains a fundamental diagnostic tool in clinical practice \cite{ref_Remmen}. 
The ability to distinguish between the stiffness of healthy and pathological tissues, such as tumors or tissue fibrosis, is a critical skill for clinicians. 
Traditionally, bedside teaching has served as the primary approach for acquiring these skills by providing medical students with direct exposure to diverse clinical cases.
However, the recruitment of patients with specific pathologies for the training of physical examination is frequently constrained by potential health risks, ethical concerns, and the inherent instability of the patients' condition \cite{ref_Krautter}.
Therefore, medical educators also conduct medical training relying on static manikins, but these models fail to replicate the dynamic range of human tissue stiffness.

Aiming to deliver these mechanical variations, several haptic technologies have been explored. 
Active robotic systems, such as motor-driven kinesthetic interfaces, are widely used for the integration with virtual environments to simulate palpation scenarios \cite{ref_Ullrich}\cite{ref_Hergenhan}. 
Although such systems can provide compelling force output, they are often limited by their bulk and don't support multi-finger interaction, which reduces the localized tactile fidelity. 
To transition from point-based virtual interactions toward more realistic physical simulations, recent research has emphasized the development of tangible tactile displays that support multi-finger interaction.
Passive mechanisms based on smart materials offer advantages in terms of compactness and variable stiffness, such as magnetorheological (MR) fluids \cite{ref_Qin}. 
Currently, some studies have used MR fluids to simulate tissue palpation, where the stiffness is increased through magnetically induced chain-like structures \cite{ref_Takano}\cite{ref_Liu}. 
However, these systems typically require complex sealing devices or high-voltage power supplies, posing safety risks and limiting their practical application.

To address these challenges, more research attention has been paid to jamming effects as a robust and inherently safe alternative for stiffness modulation. 
Within this domain, current methods for modulating stiffness primarily include granular and layer jamming.
Granular jamming utilizes the transition of loose particles into a solid-like state under vacuum pressure. 
Leveraging ground coffee inside the granular jamming chamber, Sikander et al. developed a tactile display to reproduce the feel of nodules embedded beneath the skin, allowing for variable stiffness to enhance the realism of clinical palpation \cite{ref_Sikander1}\cite{ref_Sikander2}.
Similarly, Rørvik et al. proposed a haptic surface for palpation training using ferrogranular jamming \cite{ref_Rørvik}. 
Various actuation approaches to implement particle jamming in tactile displays,  including motorized constraint, motorized compression, magnetism jamming, and vacuum jamming, were developed and tested by Brown et al \cite{ref_Brown}.
Experimental characterizations revealed that the vacuum-powered jamming interface was able to produce the most significant change in hardness and achieve the most consistent surface uniformity.
However, despite the effectiveness of granular jamming in modulating stiffness, these systems are frequently compromised by particle settling and inconsistent density, which may lead to non-uniform tactile feedback. 
Alternatively, layer jamming utilizes friction between stacked sheets to provide high structural rigidity and achieve highly consistent tactile sensations.
Several jamming mechanisms, including granular jamming with soft, fine, and rigid particles, as well as stretchable and non-stretchable layer jamming, were evaluated by He et al. for simulating muscle guarding during abdominal palpation \cite{ref_He}.
The results indicated that the non-stretchable layer jamming mechanism achieved the highest absolute stiffness, but had limitations in simulating larger and curved shapes.
Consequently, the inherent mechanical constraints of layered structures restrict their capacity for shape adaptation, posing a significant challenge for the representation of complex and irregular shapes.

Inspired by the geometry of chainmail, topologically interlocking structures have been investigated for their capacity to modulate stiffness.
This concept was initially explored through computational models, where chainmail-linked mechanisms were utilized as the bounding region to simulate the non-linear deformation of soft tissues in a virtual surgical environment \cite{ref_Zhang}. 
These computational simulations demonstrated that the mechanical interaction between discrete interlocking elements could effectively model the incompressibility and relaxation behavior of soft tissues.
Some recent studies have transitioned toward the physical implementation of structured fabrics within vacuum-driven systems. 
Wang et al. demonstrated structured fabrics with tunable bending modulus\cite{ref_Wang}.
Experimental results found that the structured fabric with small external pressure became much stiffer than in its relaxed configuration.
Furthermore, the studies of structured fabrics have demonstrated that both peak load capacity and energy absorption are significantly enhanced as the confining pressure is increased \cite{ref_Yuan}.
By encapsulating 3D-printed structured fabrics in flexible membranes, these approaches maintain geometric flexibility, while utilizing the jamming of interlocking units to achieve significant stiffness modulation.

Despite these advancements, the integration of structured fabrics into tactile displays specifically designed for medical palpation remains largely unexplored.
In this paper, a variable-stiffness and shape-conformable tactile display (VSTaI) is presented that utilizes 3D-printed structured fabrics to simulate various stiffness of soft tissues. 
The display is constructed using a tri-layer architecture: a middle layer containing a 3D-printed structured fabric within the vacuum membrane, sandwiched between two silicone membranes that mimic human tissue. 
By switching to vacuum status, the interlocking elements of the fabric are jammed together, allowing the display to transition from a highly compliant state to a rigid one. 
Four geometric patterns of units were designed, fabricated, and experimentally characterized to compare these fixed pattern-geometry combinations and to establish the stiffness design space achievable across the fabricated samples.
The experimental results are further compared with physiological tissue values to establish a baseline for the stiffness parameters required to simulate various pathological conditions in tactile training systems.

\section{Design and Fabrication}
To evaluate the influence of structural topology and scale on stiffness modulation, four patterns with different geometric parameters were selected for design and fabrication.

\subsection{Structured Fabrics}
\begin{figure}[]
\centering
\subfloat[Circular ring]{
\includegraphics[width=0.17\textwidth]{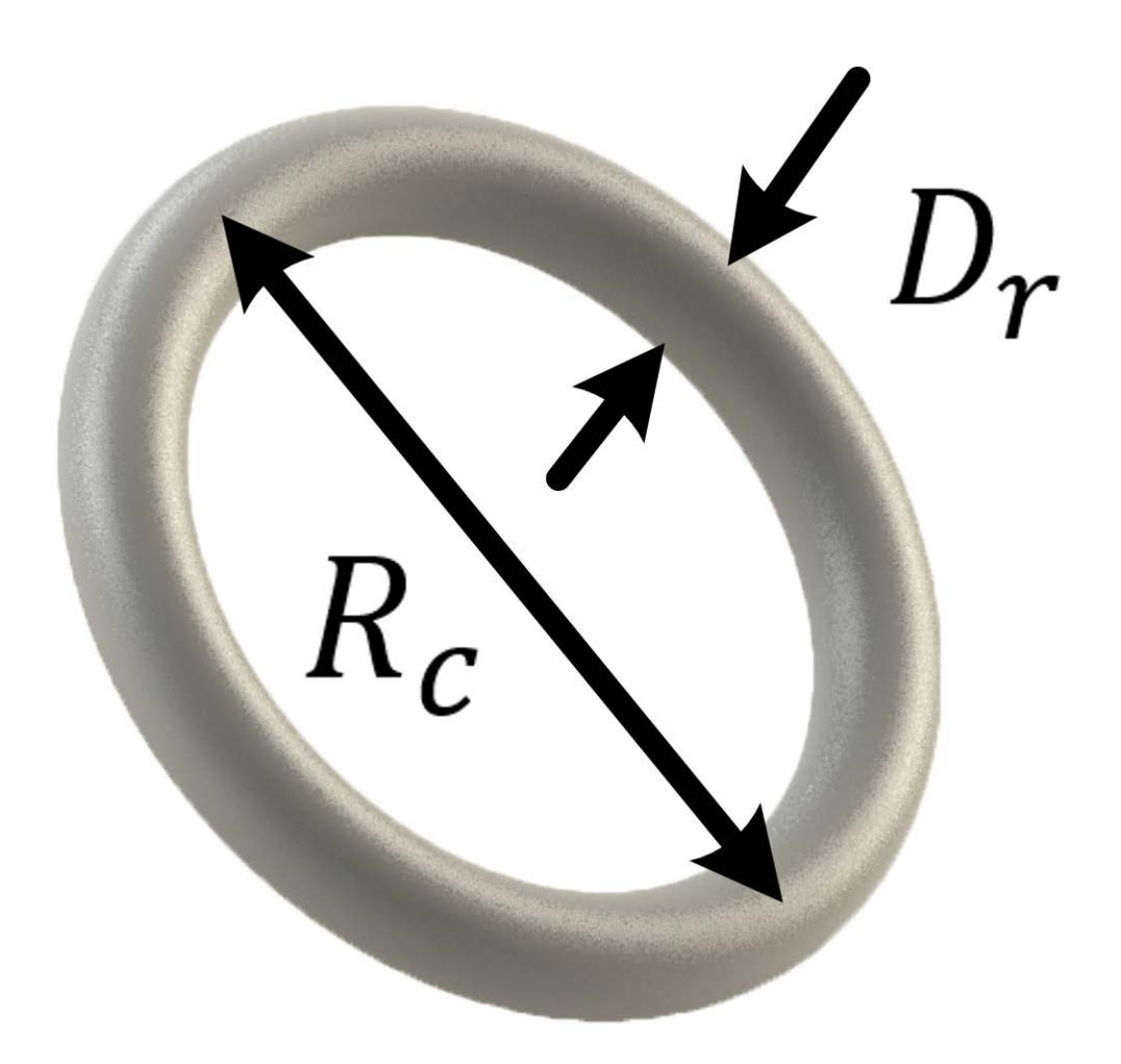}
\label{fig_circleCell}} ~~~~~
\subfloat[4-in-1 circular]{
\includegraphics[width=0.2\textwidth]{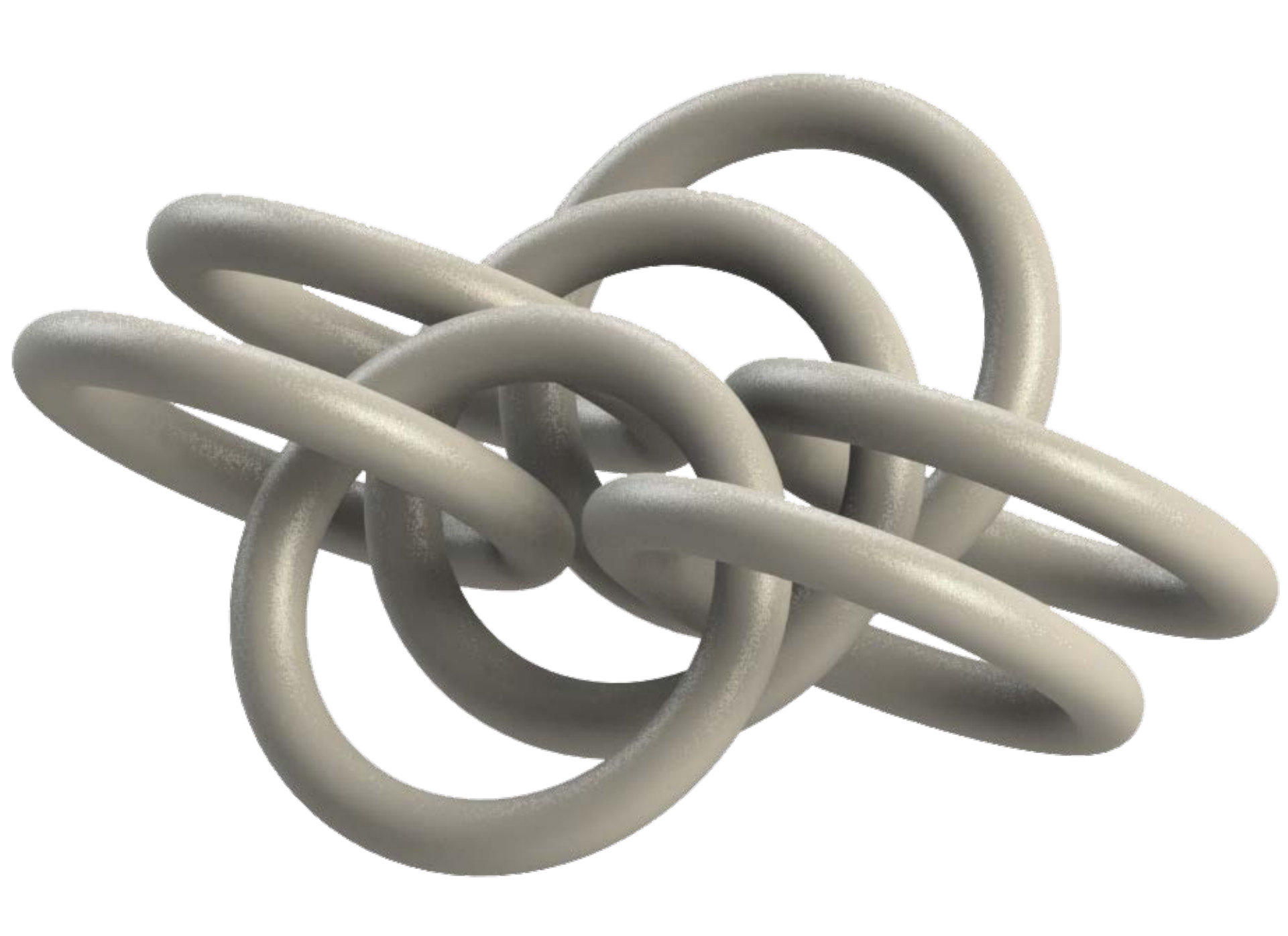}
\label{fig_4in1Cells}} ~~~
\subfloat[6-in-1 circular]{
\includegraphics[width=0.21\textwidth]{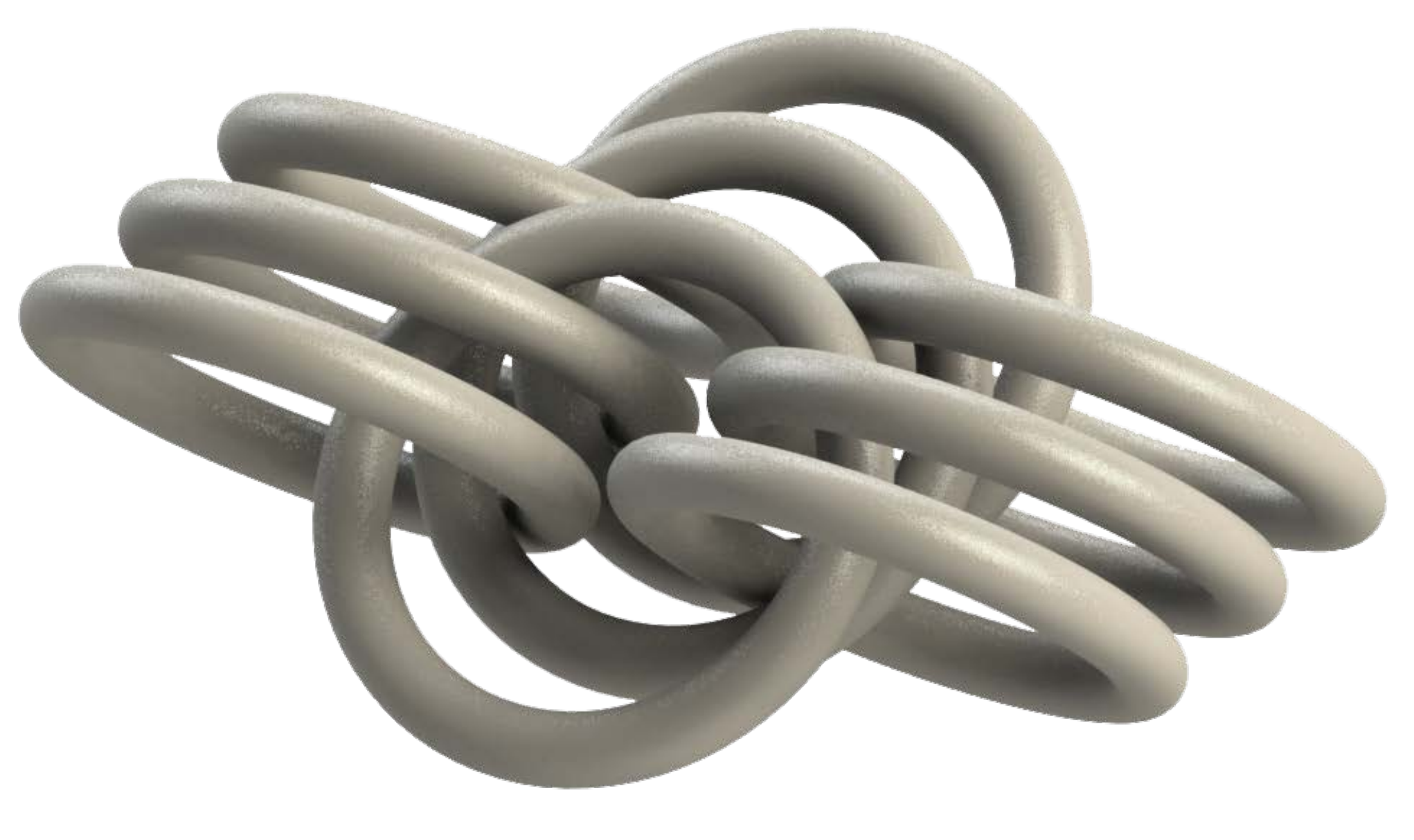}
\label{fig_6in1Cells}}\\
\subfloat[Octa.]{
\includegraphics[width=0.17\textwidth]{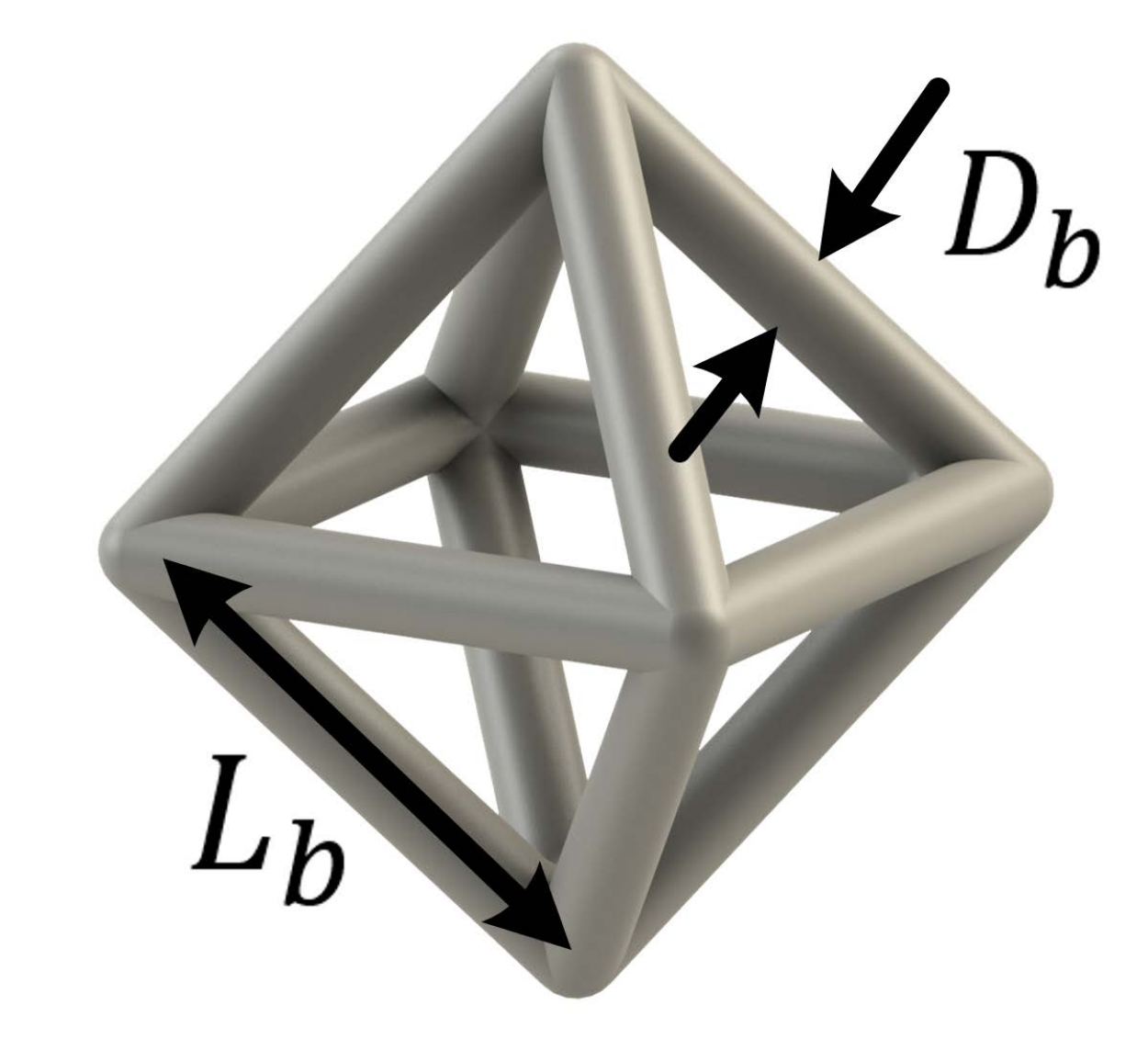}
\label{fig_octCell}}~~~
\subfloat[Octa. fabrics]{
\includegraphics[width=0.2\textwidth]{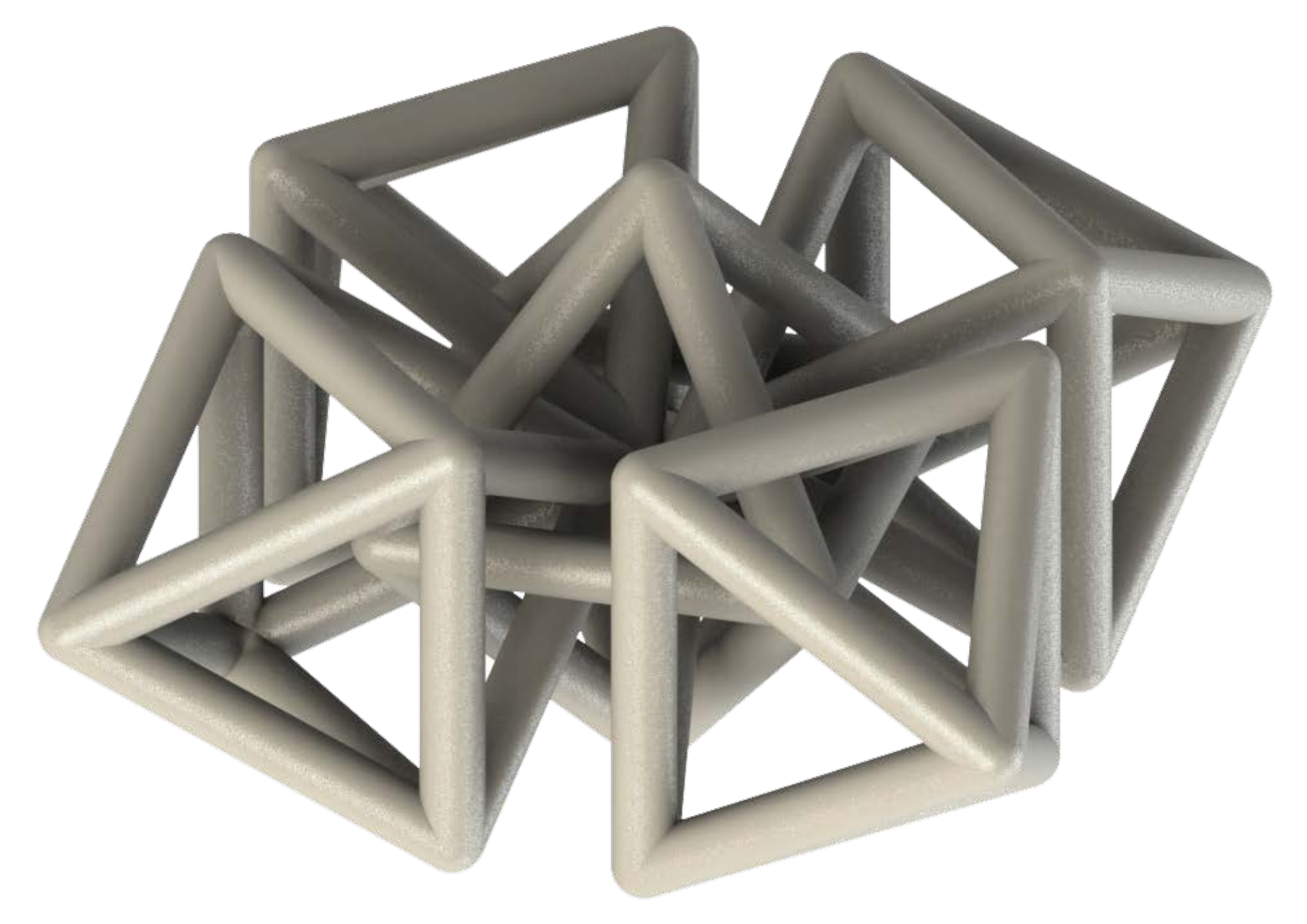}
\label{fig_octCells}}~~~
\subfloat[Hex-bipy.]{
\includegraphics[width=0.21\textwidth]{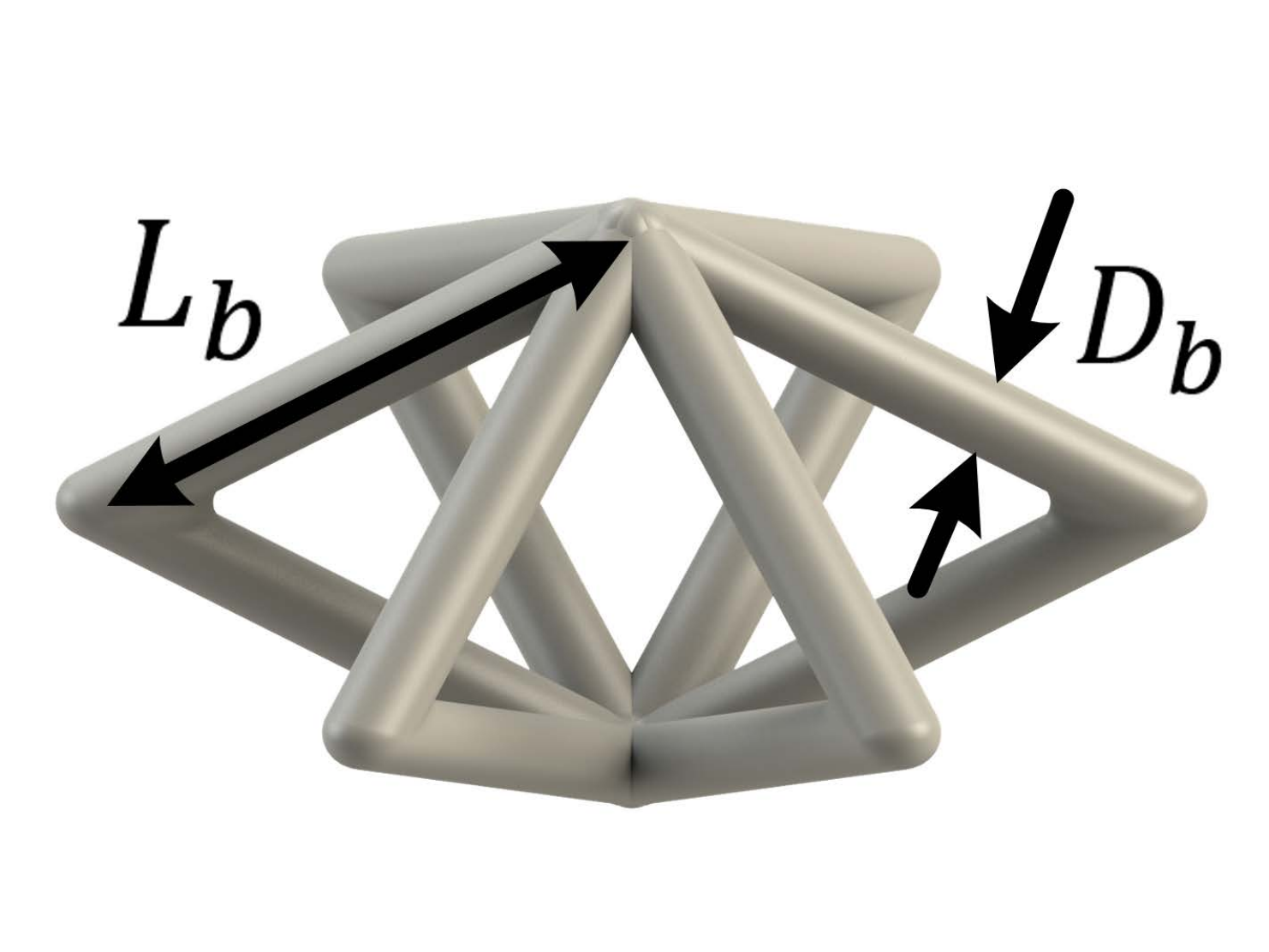}
\label{fig_hexCell}} 
\subfloat[Hex-bipy. fabrics]{
\includegraphics[width=0.22\textwidth]{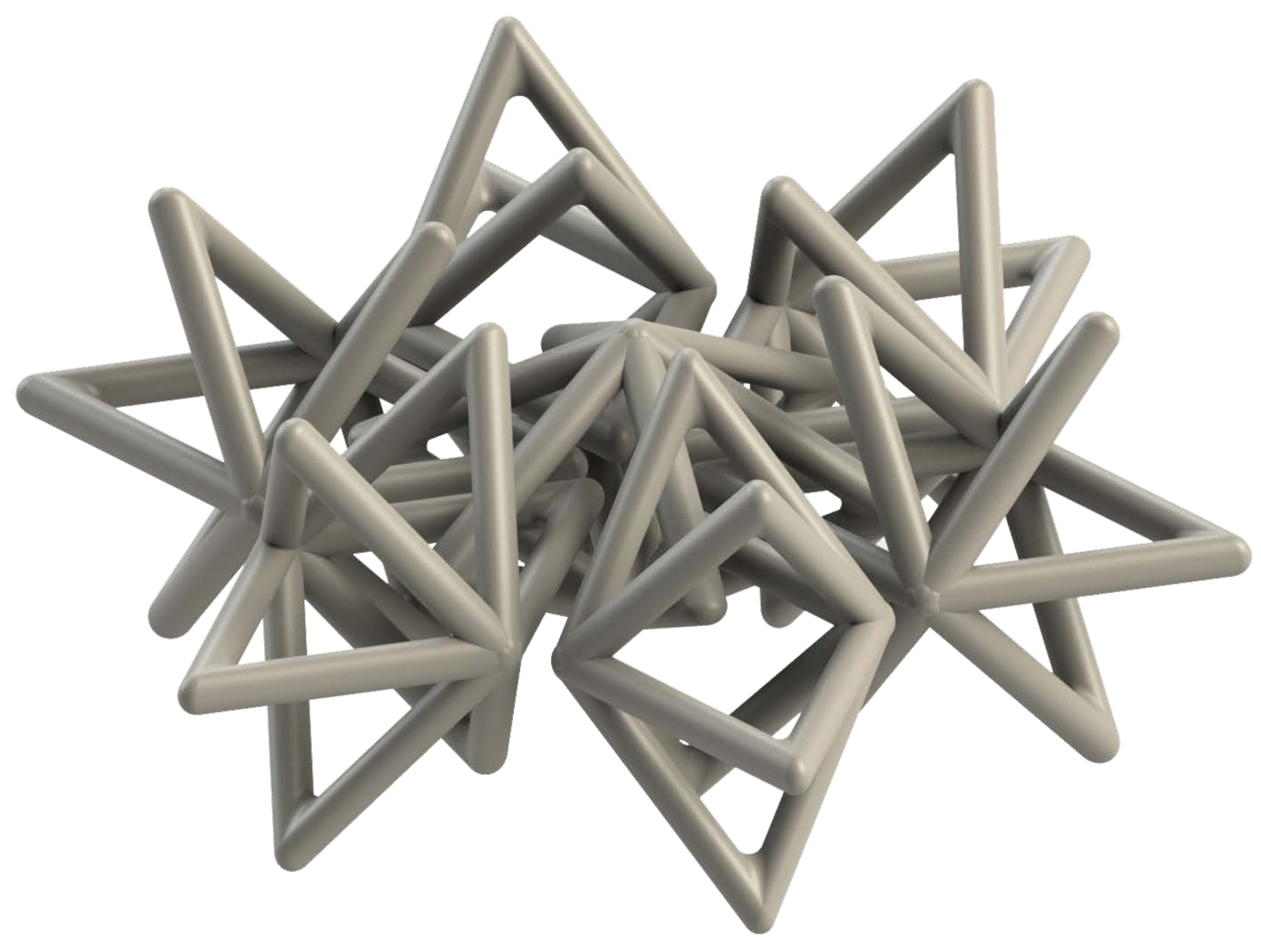}
\label{fig_hexCells}} 
\caption{Pattern types and geometric parameters. "Octa.": octahedral structure; "Hex-bipy": hexagonal bipyramid configuration without borders.}
\label{fig_patterns}
\end{figure}

As shown in Fig.~\ref{fig_patterns}, four patterns were selected for comparison: 4-in-1 circular chainmail, 6-in-1 circular chainmail, an octahedral structure, and a hexagonal bipyramid configuration without borders.

\begin{figure}[t!]
\centering
\subfloat[4-in-1 circular]{
\includegraphics[width=0.22\textwidth]{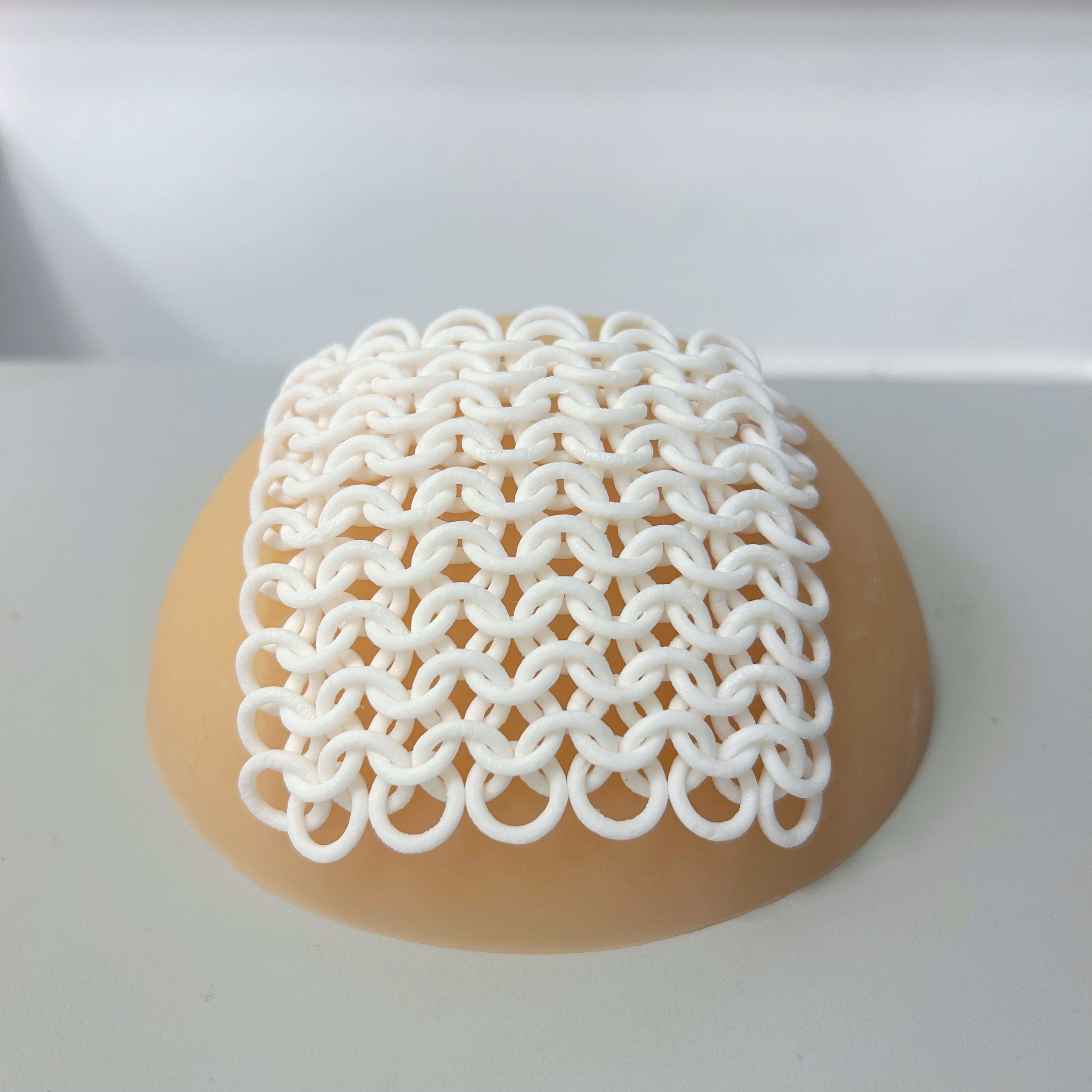}
\label{fig_4in1Print}} 
\subfloat[6-in-1 circular]{
\includegraphics[width=0.22\textwidth]{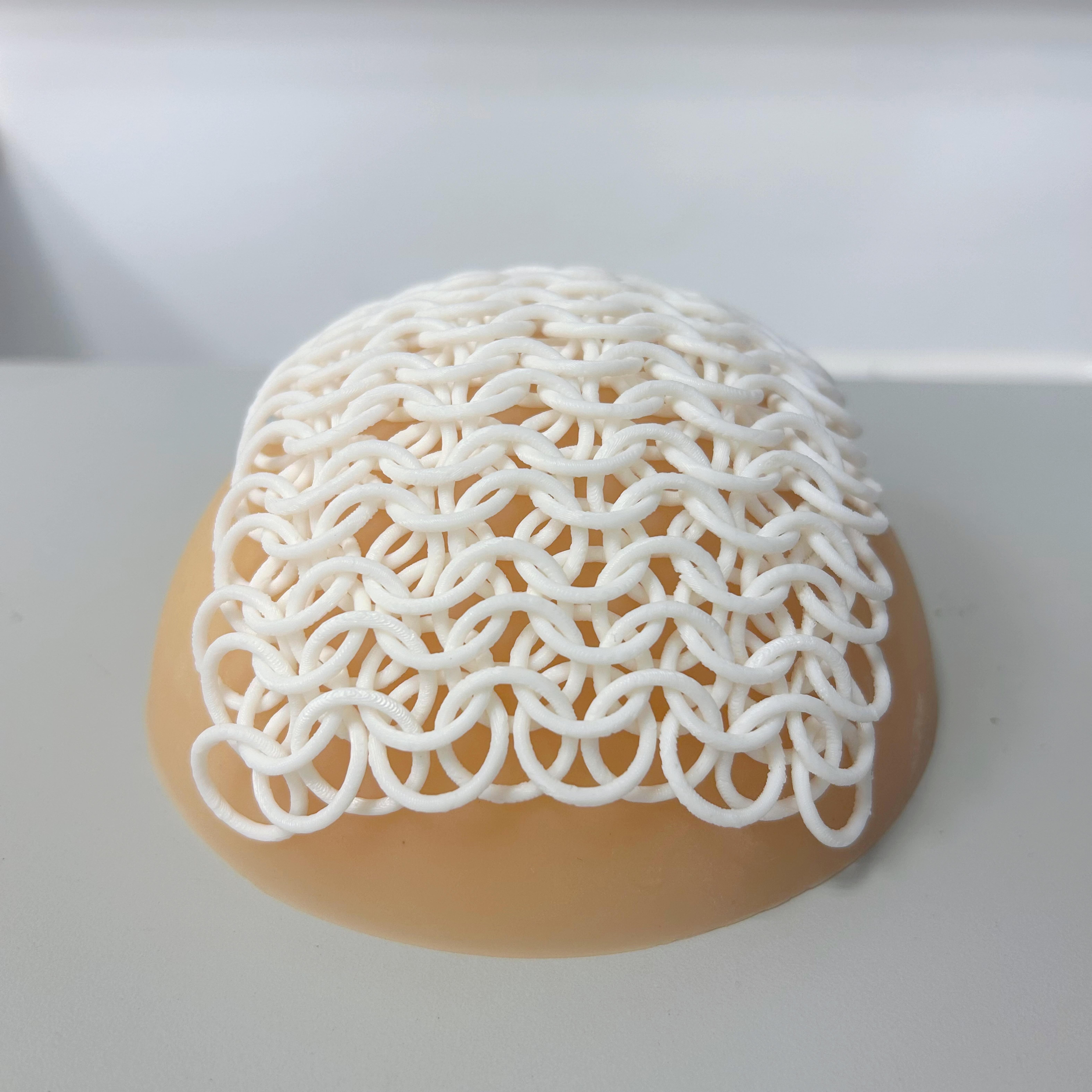}
\label{fig_6in1Print}} 
\subfloat[Octa.]{
\includegraphics[width=0.22\textwidth]{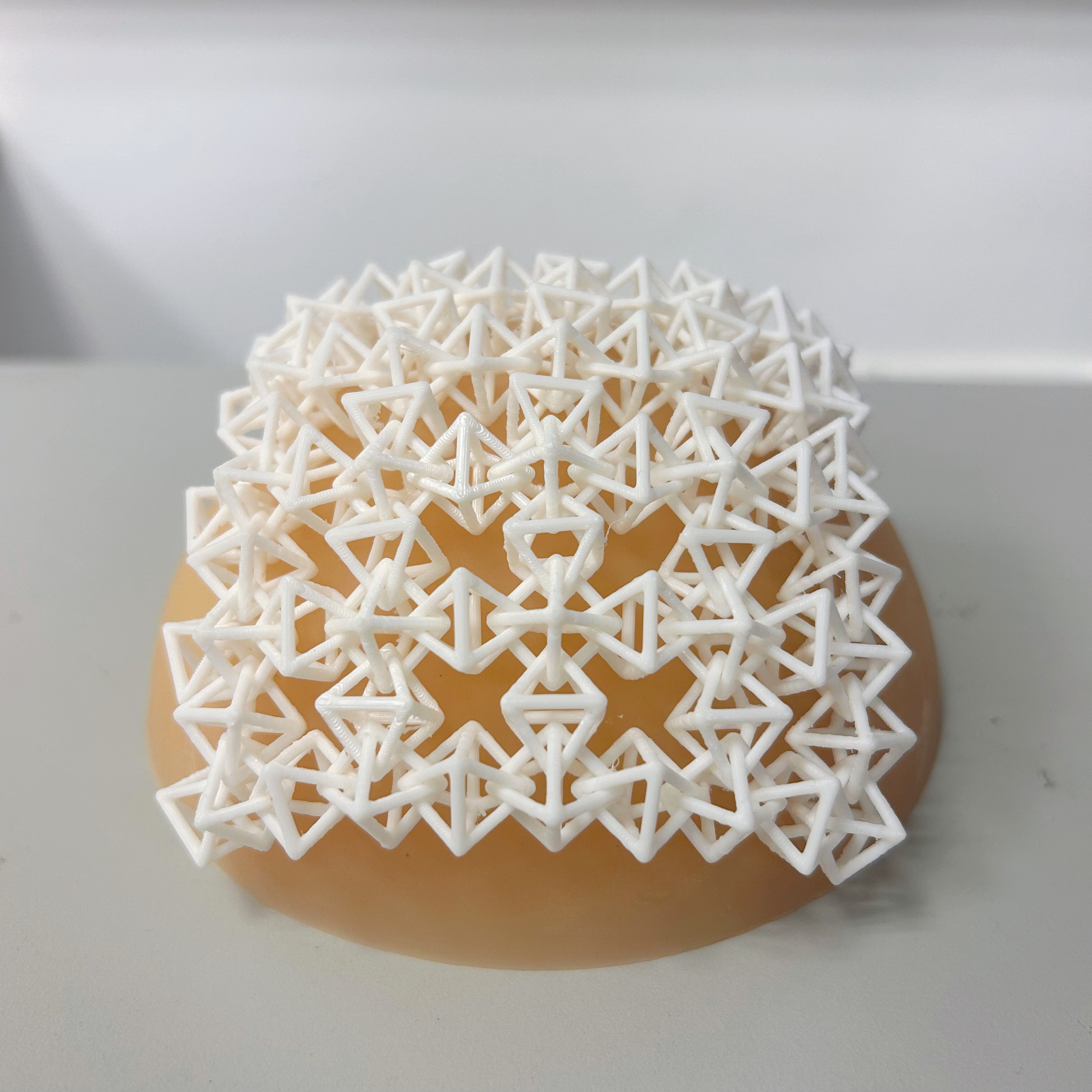}
\label{fig_octPrint}}
\subfloat[Hex-bipy.]{
\includegraphics[width=0.22\textwidth]{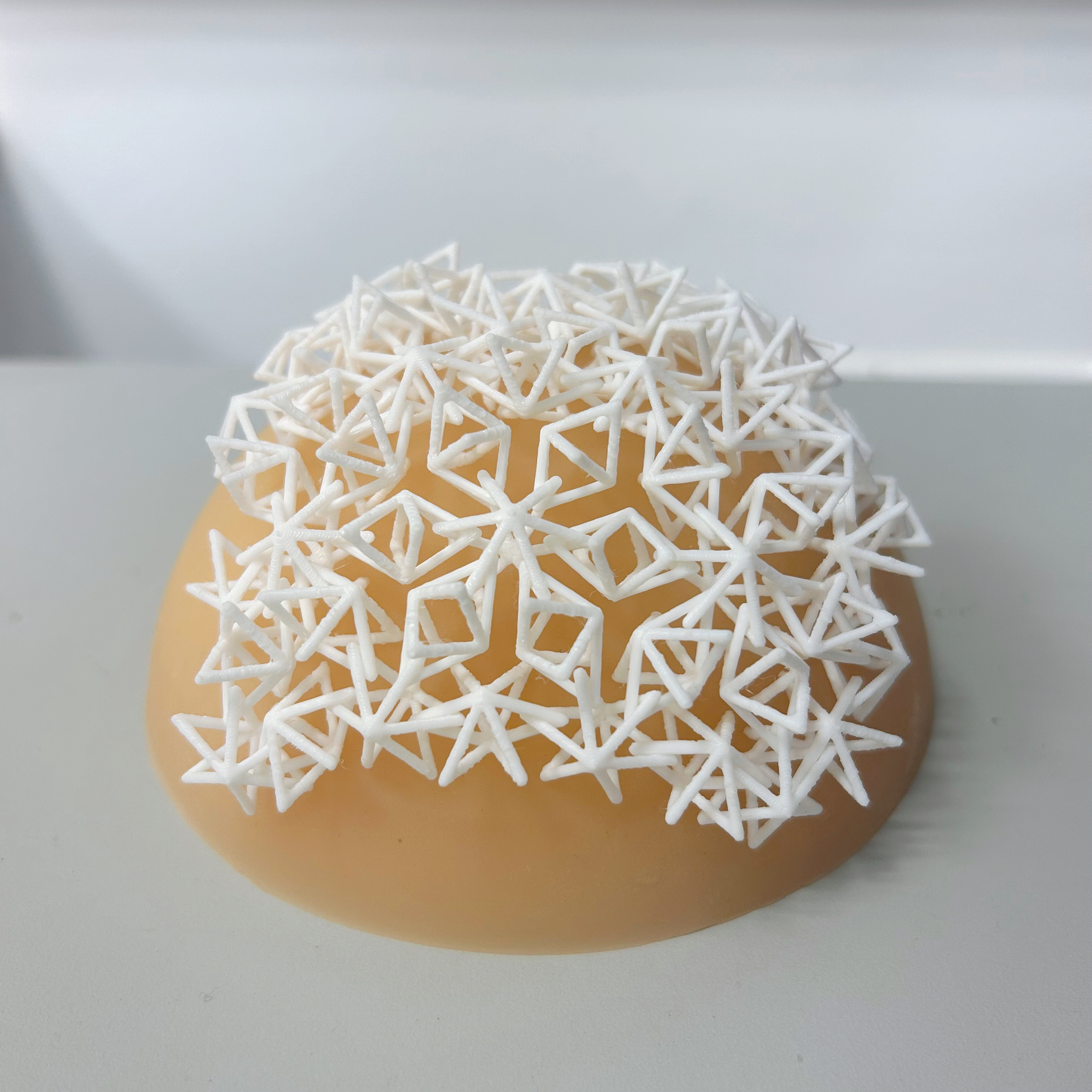}
\label{fig_hexPrint}}
\caption{3D-printed structured fabrics with various patterns.}
\label{fig_patternPrints}
\end{figure}

The 4-in-1 and 6-in-1 chainmail configurations are designed based on distinct topological arrangements of the circular unit illustrated in Fig.~\ref{fig_circleCell}, where the primary distinction resides in the interconnectivity and density.
As demonstrated in Fig.~\ref{fig_4in1Cells} and Fig.~\ref{fig_6in1Cells}, each individual internal ring of the 4-in-1 configuration is linked to four adjacent neighbors, whereas the 6-in-1 structure increases this connectivity to six neighbors per ring. 
To ensure manufacturability, two sets of geometric parameters were defined, featuring rod diameters ($D_r$) of 1.5 mm and 2.0 mm with corresponding circle radii ($R_c$) of 10 mm and 15 mm, respectively.
As shown in Fig.~\ref{fig_octCell}, the octahedral pattern is characterized by its 90-degree rotational symmetry. 
Beam diameters ($D_b$) of 1.5 mm and 2.0 mm were utilized alongside corresponding beam lengths ($L_b$) of 10 mm and 15 mm, respectively.
As demonstrated in Fig.~\ref{fig_octCells}, the fabrics based on the octahedral pattern are generated by rotating neighboring particles 90 degrees relative to one another, allowing all elements to be topologically interlocked.
For the hexagonal bipyramid configuration shown in Fig.~\ref{fig_hexCell}, the element is obtained by replicating and rotating two triangular trusses with angles of 60 degrees.
The geometric parameters were defined by beam diameters ($D_b$) of 1.5 mm and 2.0 mm, which were paired with corresponding lengths ($L_b$) of 10 mm and 15 mm, respectively.
Each truss is linked to a corresponding unit in an orthogonal orientation, which results in a structured fabric characterized by a dense interlocking network as demonstrated in Fig.\ref{fig_hexCells}.

The structured fabrics were initially designed in SolidWorks and subsequently 3D-printed on X1 Carbon (Bambu Lab). 
Polylactic acid (PLA) was utilized as the primary material, while polyvinyl alcohol (PVA) was employed as a sacrificial support to facilitate the production of the interlocking geometry. 
Following fabrication, each sample with specific parameters was trimmed to a square configuration with a side length of approximately 8 cm.
A total of eight samples were produced to include all combinations of the four structural patterns and two sets of geometric parameters. 
The flexibility of these fabrics is demonstrated in Fig.~\ref{fig_patternPrints}, which presents four representative samples conforming to a semi-circular surface to show their shape adaptability.

\subsection{Components of VSTaI}

\begin{figure}[!b]
\centering
\subfloat[Assembly]{
\includegraphics[width=0.55\textwidth]{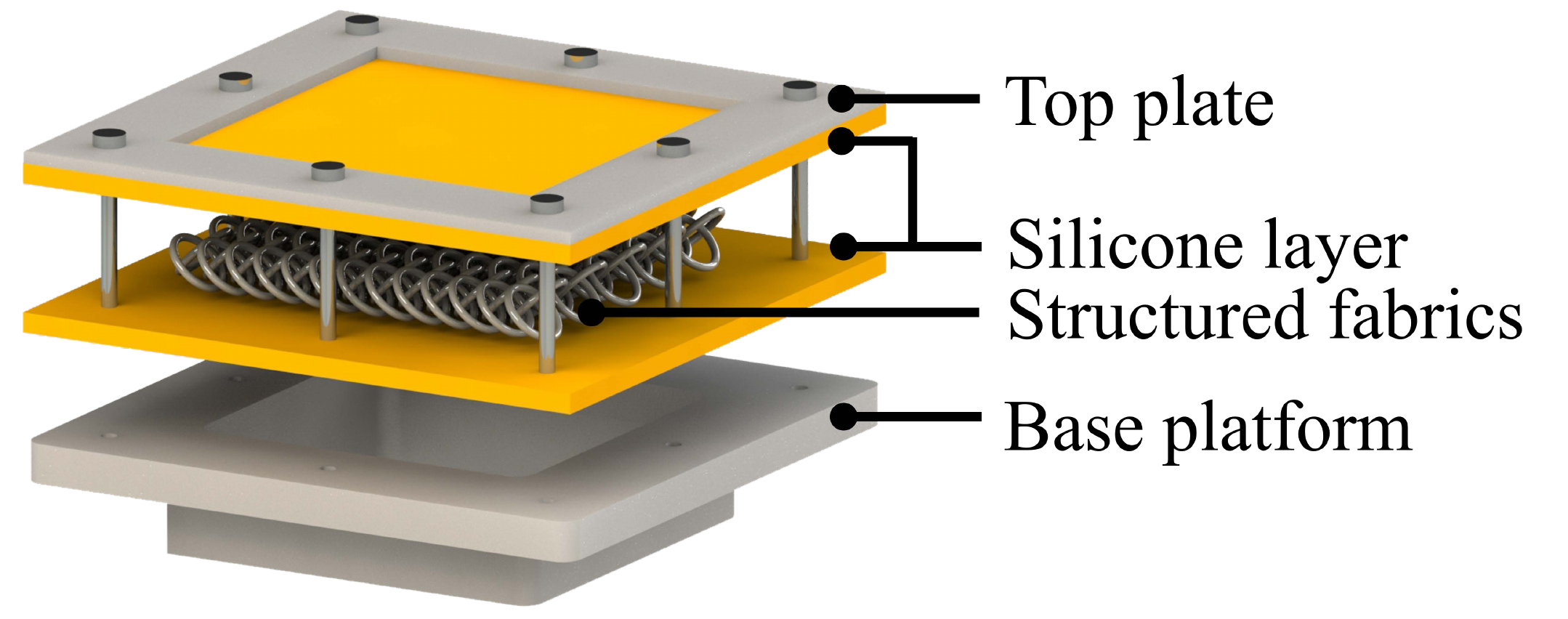}
\label{fig_assembly}} 
\subfloat[Prototype]{
\includegraphics[width=0.4\textwidth]{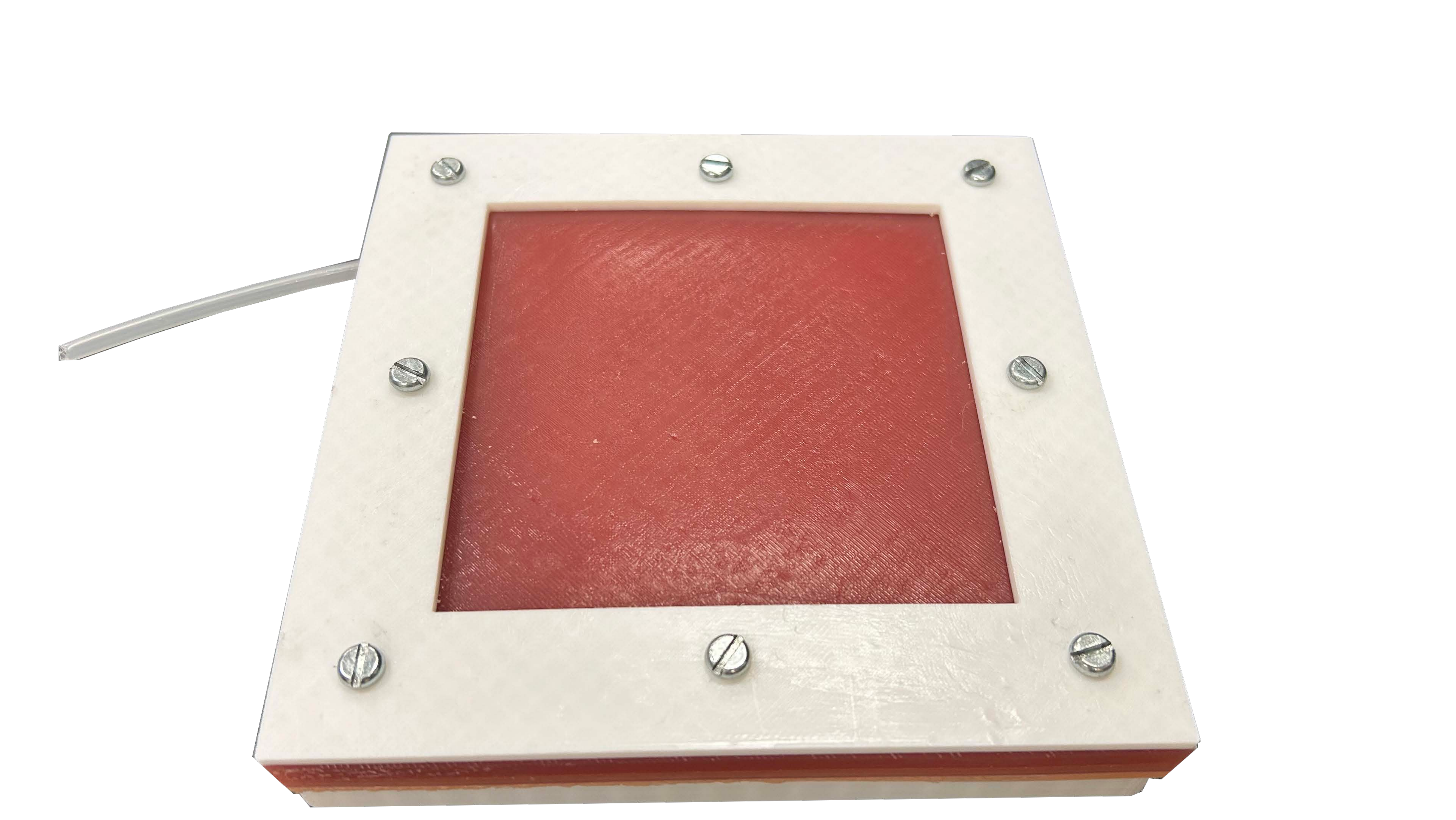}
\label{fig_vstai}} 
\caption{Assembly and prototype of VSTaI.} 
\label{fig_prototype}
\end{figure}

Based on the 3D-printed structured fabrics, VSTaI was developed as a haptic display to render different soft tissue stiffness levels. 
This prototype also serves as a platform for the characterization of stiffness modulation across various interlocking patterns and geometric parameters. 
To create a tissue-like tactile surface, the structured fabric was encapsulated within a vacuum-sealed polyethylene membrane and positioned between two 5-mm-thick layers of silicone.
These layers were molded from Ecoflex 00-30 to provide a compliant interface similar to human tissue. 
This separation of layers was intended to decouple the sealing function from the tissue-contact function.
The resulting tri-layer structure was mounted onto a platform featuring an $8 \times 8$ cm chamber. 
A rigid top plate was utilized to secure the assembly, with screws passing through the plate, the silicone layers, and the platform base to ensure that the tri-layer structure can be clamped stably.
This prototype allows the mechanical response of each fabric pattern to be precisely measured for subsequent comparison with the physiological stiffness of actual human tissues.

\section{Stiffness Characterization}

The tunable stiffness range of VSTaI was evaluated through force-indentation bench tests.
By measuring these stiffness properties under both atmospheric and vacuum conditions, the achievable stiffness range for each fabric was determined. 
This characterization provided a quantitative basis for understanding how specific fabric geometries influence the stiffness tunability with vacuum-induced jamming.

\subsection{Experimental Setup}

\begin{figure}[!b]
\centering
\subfloat[Setup]{
\includegraphics[width=0.55\textwidth]{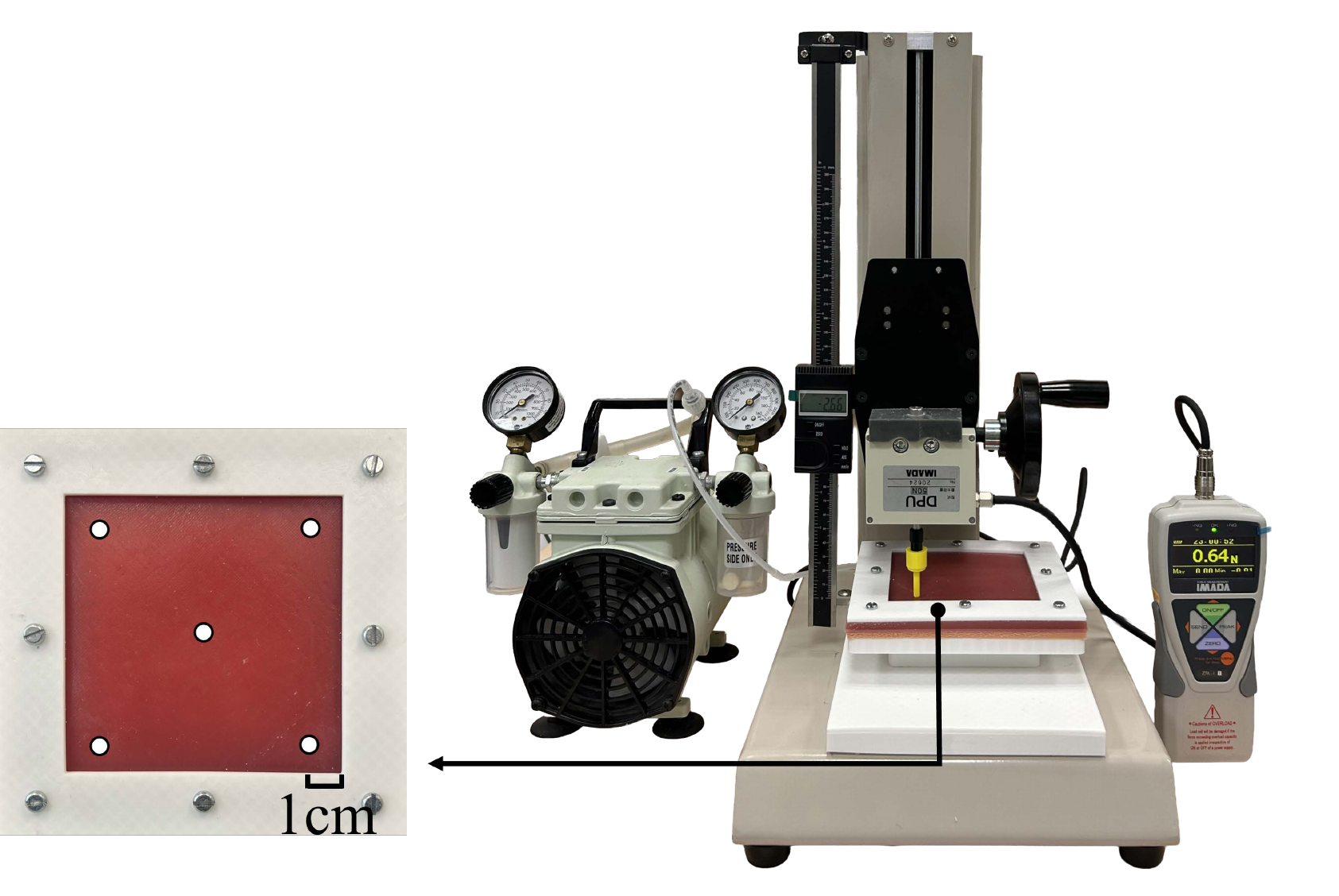}
\label{fig_testSetup}} 
\subfloat[Illustration]{
\includegraphics[width=0.23\textwidth]{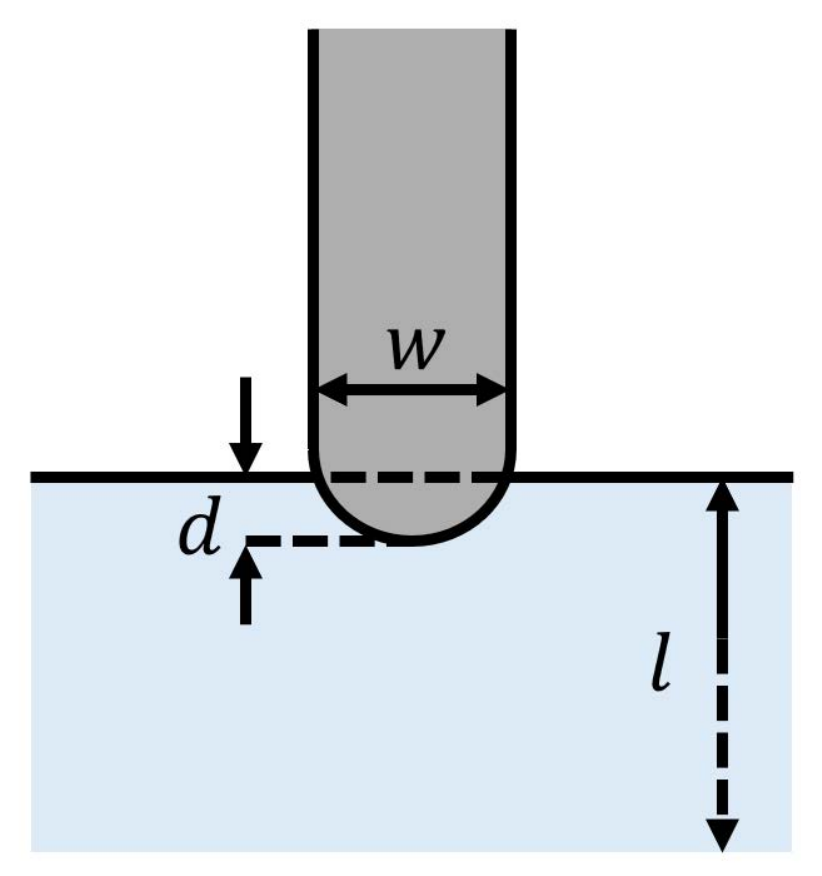}
\label{fig_stiffFig}} 
\caption{Experimental setups for stiffness characterization and illustration of stiffness calculation. } 
\label{fig_stiffTests}
\end{figure}

As shown in Fig.~\ref{fig_testSetup}, the experimental setup includes VSTaI, a regulated vacuum pump, a force test stand with a caliper, a one-way valve, and a force gauge (ZTS-DPU-50N, IMADA CO.,LTD.).
Force measurements were recorded using the force gauge mounted on the vertical carriage of the test stand.
An indenter with a 4.5 mm round tip was positioned in initial contact with the surface of VSTaI.
By operating the manual handwheel, the indenter pressed on the display, allowing for the simultaneous acquisition of force and displacement data.
To evaluate the spatial uniformity of the display, measurements were conducted at five points. 
These points included the geometric center and the four corners, with each peripheral measurement taken at a distance of 1 cm from the bezel.
The vacuum status was controlled via a vacuum pump equipped with an air pressure gauge. 
The haptic display was connected to this unit through tubing and a one-way valve, ensuring an airtight seal for the vacuum membrane. 

The stiffness of the display was calculated as Young’s modulus and quantified by performing a linear regression on the stress-strain data as illustrated in Fig.~\ref{fig_stiffFig}. 
The stiffness can be derived as follows:

\begin{equation}
\label{equa_stiff}
\begin{aligned}
E=\frac{\sigma}{\epsilon}=\frac{F/A}{d/ l}
\end{aligned}
\end{equation}

The stiffness of the model, $E$, is derived from the linear relationship between stress ($\sigma$) and strain ($\epsilon$). 
$\sigma$ is defined as the ratio of the applied force $F$ to the contact area $A$, while $\epsilon$ is determined by the indentation depth $d$ relative to the initial vertical span $l$.
Considering the hemispherical geometry of the indenter tip with diameter $w$, the contact area $A$ can be expressed as a function of the indentation depth. 
If $d < w/2$, the contact area is calculated as:

\begin{equation}
\label{equa_surface}
\begin{aligned}
A = \pi (wd - d^2)
\end{aligned}
\end{equation}

For indentation depths where $d \geq w/2$, the contact reaches the equator of the hemisphere and the projected contact area $A$ becomes constant:

\begin{equation}
\label{equa_surface}
\begin{aligned}
A = \pi (w/2)^2
\end{aligned}
\end{equation}

\begin{table}[t!]
\centering
\caption{Classification of geometric parameter groups for the tested structural patterns.\label{tab_geoGroups}}
\begin{tabular}{lll}
\hline\hline 
Patterns & ~~~~4-in-1 circular & ~~~~6-in-1 circular     \\ \hline
G1       & ~~~~$D_r$=1.5mm, $R_c$=10mm     & ~~~~$D_r$=1.5mm, $R_c$=10mm                        \\
G2       & ~~~~$D_r$=2mm, $R_c$=15mm       & ~~~~$D_r$=2mm, $R_c$=15mm                \\ \hline
Patterns & ~~~~Octahedral      & ~~~~Hexagonal bipyramid \\ \hline
G1   &    ~~~~$D_b$=1.5mm, $L_b$=10mm        & ~~~~$D_b$=1.5mm, $L_b$=10mm                 \\
G2       & ~~~~$D_b$=2mm, $L_b$=15mm         & ~~~~$D_b$=2mm, $L_b$=15mm                 \\ \hline \hline
\end{tabular}
\end{table}

Force-displacement measurements were conducted under both atmospheric and vacuum conditions. 
For the vacuum trials, a negative pressure of 90 kPa was applied and allowed to stabilize for 30 seconds before recording the data. 
The experimental trials consisted of eight different samples, representing four structural patterns with two parametric configurations each, as summarized in Table~\ref{tab_geoGroups}. 
By evaluating two pressure conditions across five measurement points per sample, a total of 80 trials were generated. 
For each trial, the indentation depth was measured at 0.5 mm intervals until the maximum displacement reached 10 mm.
This value was selected to provide a consistent and measurable deformation range for the comparative mechanical characterization of the tested samples.

\subsection{Stiffness Characterization Results}

\begin{figure}[b!]
\centering
\subfloat[4-in-1 circular]{
\includegraphics[width=0.5\textwidth]{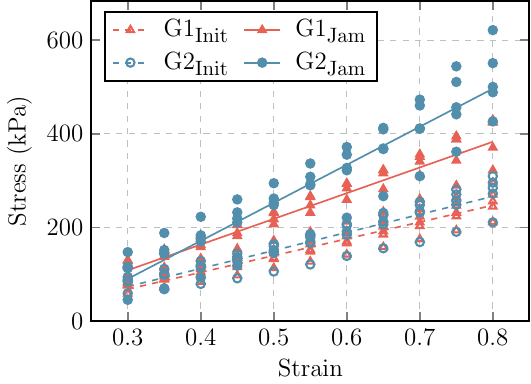}
\label{fig_LinePlot4}} 
\subfloat[6-in-1 circular]{
\includegraphics[width=0.5\textwidth]{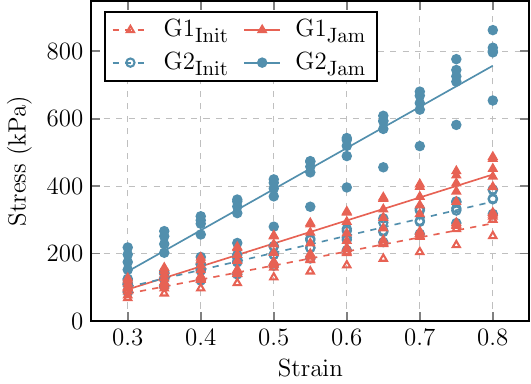}
\label{fig_LinePlot6}} \\
\subfloat[Octahedral]{
\includegraphics[width=0.5\textwidth]{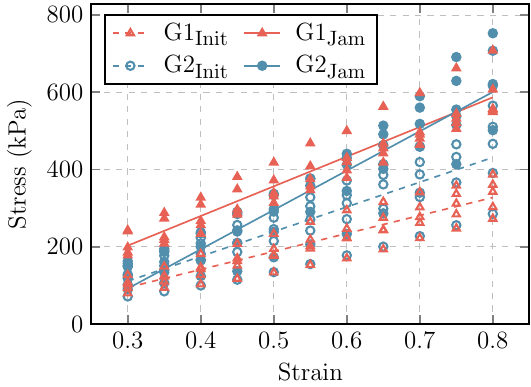}
\label{fig_LinePlotO}} 
\subfloat[Hexagonal bipyramid]{
\includegraphics[width=0.5\textwidth]{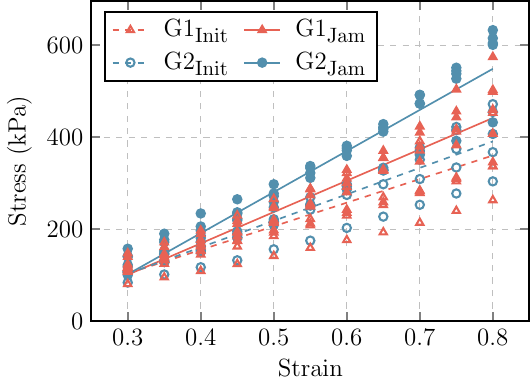}
\label{fig_LinePlotH}} \\
\caption{Stress-strain relationships and linear stiffness fitting for four geometric patterns under vacuum and non-vacuum conditions. 
 “Init” represents initially atmospheric pressure; "Jam" represents vacuum jamming pressure.
(a) $G1_{Init}: R^2=0.993$, $G1_{Jam}: R^2=0.994$, $G2_{Init}: R^2=0.993$, $G2_{Jam}: R^2=0.989$. (b) $G1_{Init}: R^2=0.993$, $G1_{Jam}: R^2=0.992$, $G2_{Init}: R^2=0.995$, $G2_{Jam}: R^2=0.992$. (c) $G1_{Init}: R^2=0.992$, $G1_{Jam}: R^2=0.997$, $G2_{Init}: R^2=0.992$, $G2_{Jam}: R^2=0.973$. (d) $G1_{Init}: R^2=0.994$, $G1_{Jam}: R^2=0.980$, $G2_{Init}: R^2=0.993$, $G2_{Jam}: R^2=0.987$.
}
\label{fig_LinePlots}
\end{figure}

To ensure a standardized characterization of the device, the data analysis was focused on the specific strain range between 0.3 and 0.8. 
This interval was selected as a practical fitting window for comparative stiffness estimation. 
The lower-strain region was excluded because it is more sensitive to initial contact establishment and local rearrangement of the interlocking units, whereas the higher-strain end was excluded to reduce the influence of stronger geometric nonlinearity and boundary effects in the finite-thickness layered structure. 
Within this interval, the stress-strain response was approximately linear under the tested conditions.
Therefore, the stiffness for each individual trial was determined by calculating the slope of the stress-strain curve using a least-squares linear regression. 
The characteristic stiffness for each structural configuration was subsequently established as the arithmetic mean of these individual slopes. 

\begin{figure}[]
\centering
\includegraphics[width=1\textwidth]{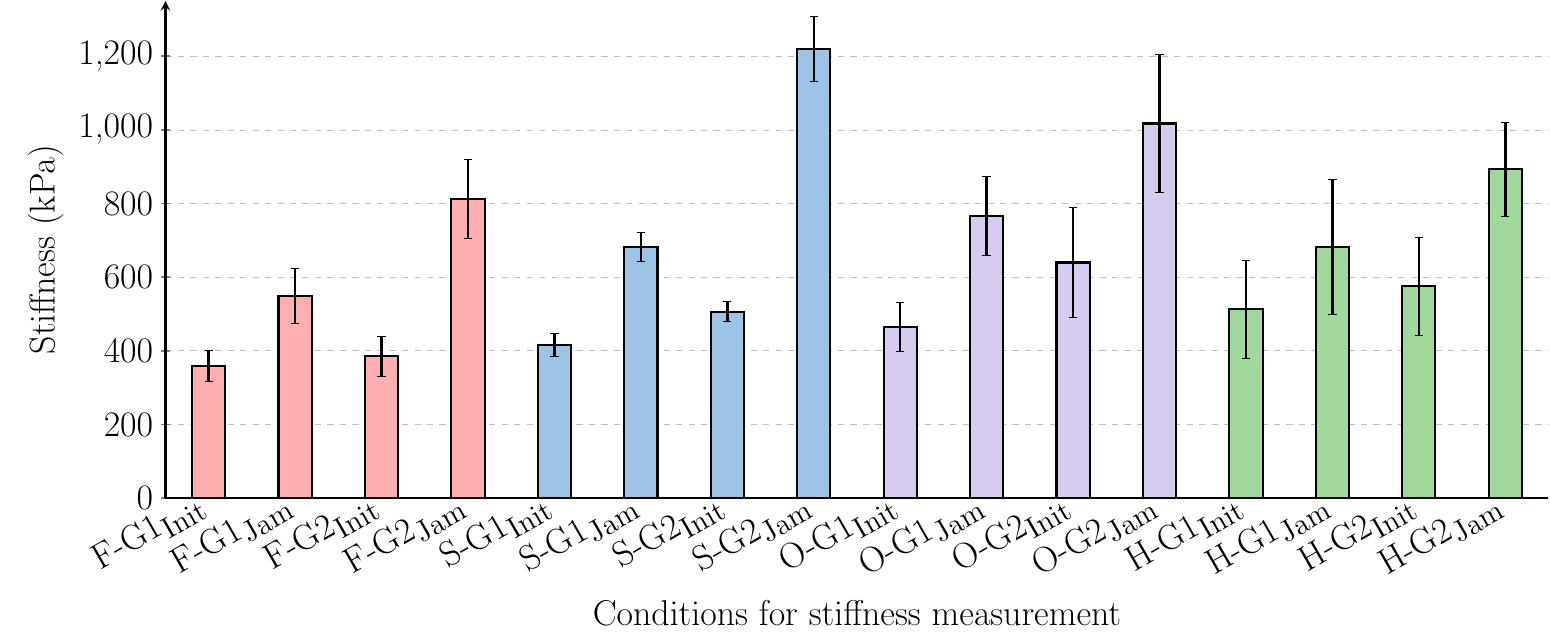}
\caption{Average and standard deviation of stiffness in each condition. F- represents 4-in-1 circular; S- represents 6-in-1 circular; O- represents Octahedral; H- represents Hexagonal bipyramid.} 
\label{fig_stiffResults}
\end{figure}

For simplicity, these groups are denoted as G1 and G2 for each pattern as demonstrated in Table~\ref{tab_geoGroups}. 
The stress-strain relationships obtained from bench testing are illustrated in Fig.~\ref{fig_LinePlots}. 
Data for G1 and G2 are represented by triangular and circular markers, respectively. 
The linear fit for each experimental condition is represented by the corresponding dashed and solid lines, denoting the initial atmospheric state and the vacuum jamming state, respectively.
For all configurations, the captured data points and corresponding linear fit exhibit a consistent upward shift in slope when transitioning from the initial atmospheric state to the vacuum jamming state. 
It is observed that the G2 geometry group consistently achieves higher stress values for a given strain compared to G1 across all patterns. 
Notably, the 6-in-1 circular pattern and the hexagonal bipyramid structure demonstrate a more significant difference between the G1 and G2 results, suggesting that these geometries are particularly sensitive to structural parameter variations.

The calculated average stiffness values, derived from the mean of five measurement points per condition, are summarized in Fig~\ref{fig_stiffResults}. 
The results reveal that vacuum jamming significantly enhances the stiffness of the display. 
The $\mathrm{S-G2_{Jam}}$ configuration reached the maximum mean stiffness of $1220.05 \text{kPa}$ under vacuum jamming, representing a $140\%$ increase from its initial state of $506.62\text{ kPa}$. 
The $\mathrm{H-G1}$ configuration demonstrated the most limited change in stiffness, with values ranging from $513.09\text{ kPa}$ to $681.78\text{ kPa}$ between the atmospheric and jammed states.
Higher standard deviations were observed in the hexagonal and octahedral configurations.
The architectural design of these patterns, consisting of sharp vertices and rigid rods, undergoes mechanical interlocking under vacuum pressure. 
The small shifts in their relative contact points significantly influence the global stiffness of the display. 
The measured stiffness was strongly influenced by the initial random distribution of the units, as it introduced internal gaps and led to a higher standard deviation than that observed in the circular patterns.

\begin{figure}[t!]
\centering
\subfloat[4-in-1 circular]{
\includegraphics[width=0.45\textwidth]{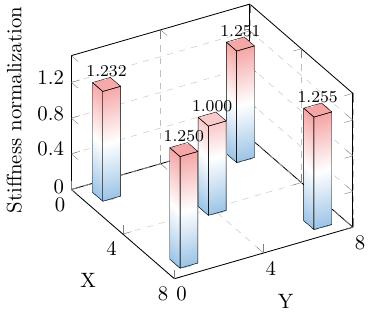}
\label{fig_5point4}} 
\subfloat[6-in-1 circular]{
\includegraphics[width=0.45\textwidth]{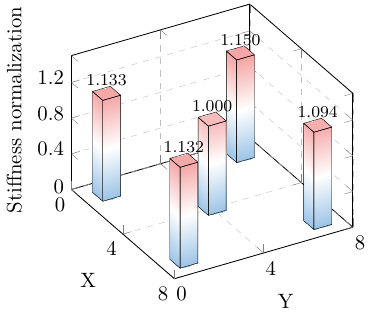}
\label{fig_5point6}} \\
\subfloat[Octahedral]{
\includegraphics[width=0.45\textwidth]{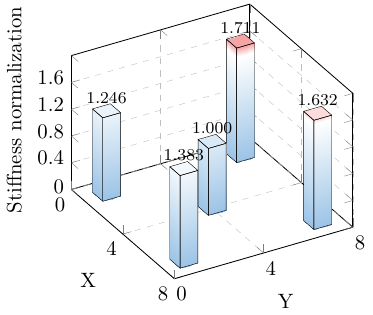}
\label{fig_5pointO}} 
\subfloat[Hexagonal bipyramid]{
\includegraphics[width=0.45\textwidth]{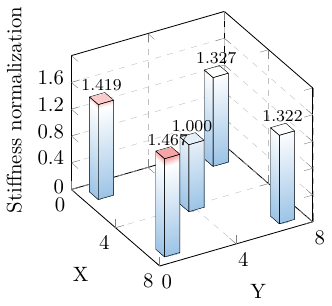}
\label{fig_5pointH}} \\
\caption{Spatial distribution of stiffness results on five measured points.} 
\label{fig_spatialPlots}
\end{figure}

To further evaluate the effects of pattern (4-in-1 circular, 6-in-1 circular, Octahedral, and Hexagonal bipyramid), geometric parameter group (G1 and G2), and air pressure condition (Init and Jam), a three-way analysis of variance (ANOVA) was performed. 
According to the Shapiro-Wilk test, except for one group (F-$G2_{Jam}$, $p=0.038$), the stiffness data in the other groups followed a normal distribution.
The three-way interaction was not significant ($F = 2.717$, $p = 0.052$), indicating that the two-way interaction effects did not differ significantly across levels of the third factor. 
Meanwhile, all two-way interactions were significant, including pattern $\times$ geometric parameter group ($F = 2.802$, $p = 0.047$), pattern $\times$ air pressure condition ($F = 4.530$, $p = 0.006$), and geometric parameter group $\times$ air pressure condition ($F = 21.598$, $p < 0.001$). 
Significant main effects were also observed for pattern ($F = 13.320$, $p < 0.001$), geometric parameter group ($F = 68.364$, $p < 0.001$), and air pressure condition ($F = 199.255$, $p < 0.001$). 
Bonferroni-corrected simple-effects analyses were then performed to examine the significant two-way interactions.

For the interaction between pattern and air pressure condition, the stiffness under vacuum jamming pressure was significantly greater than that under atmospheric pressure across all four patterns (4-in-1 circular: $p < 0.001$; 6-in-1 circular: $p < 0.001$; Octahedral: $p < 0.001$; Hexagonal bipyramid: $p < 0.001$). 
For the pattern $\times$ geometric parameter group interaction, G2 yielded significantly higher stiffness than G1 across all four patterns (4-in-1 circular: $p = 0.004$; 6-in-1 circular: $p < 0.001$; Octahedral: $p < 0.001$; Hexagonal bipyramid: $p = 0.007$).
For the interaction between geometric parameter group and air pressure condition, the stiffness under vacuum jamming pressure was found to be statistically higher than that under atmospheric pressure in both G1 ($p < 0.001$) and G2 ($p < 0.001$), with a larger jamming-induced increase observed under G2.

The stiffness values for the five measurement points under each experimental condition were calculated and normalized relative to the central sampling point to account for spatial variance. 
The average normalized stiffness was computed across the four parametric configurations for each point to characterize the uniformity of the specific fabric pattern. 
These values are compared in Fig.~\ref{fig_spatialPlots}, illustrating the normalized stiffness at five measurement locations under vacuum jamming conditions.
The results indicate that the circular patterns demonstrate better uniformity across the display surface. 
Specifically, the stiffness normalization values for the 4-in-1 and 6-in-1 circular patterns remained within relatively narrow ranges of 1.000 to 1.255 and 1.000 to 1.150, respectively. 
These results align with the lower standard deviations presented in Fig~\ref{fig_stiffResults}, suggesting a highly repeatable and stable stiffness rendering across the VSTaI surface.
However, the octahedral and hexagonal bipyramidal patterns exhibited significantly lower spatial uniformity. 
The octahedral pattern showed a peak normalization value of 1.711, while the hexagonal bipyramidal pattern reached 1.467 at the peripheral measurement locations.

\section{Discussion}
According to previous research \cite{ref_Guimarães} \cite{ref_Singh}, human tissues span a wide range of stiffness, with elastic moduli ranging from $11\text{ Pa}$ for mucus to $20\text{ GPa}$ for cortical bone. 
Within this range, organs and skin are classified as soft tissues, with moduli typically in the sub-megapascal range.
The experimental results indicate that the VSTaI with the specific patterns and geometries achieved an effective stiffness range between $358.24\text{ kPa}$ and $1220.05\text{ kPa}$. 
This range overlaps with much of the soft tissue and transitions into the stiffer categories, such as cartilage. 
By modulating the vacuum pressure, the display can simulate the transition from healthy, compliant abdominal regions, to higher resistance, which is typically associated with pathologies or hardened masses.
Furthermore, the choice of 3D-printed structural patterns provides a practical method for applying the interface to specific clinical training scenarios. 
It should be noted that the overall stiffness range reported in this study reflects the stiffness results obtained across multiple structured-fabric configurations rather than the real-time tuning range of a single prototype. 
Therefore, each pattern-geometry combination should be regarded as an independent hardware configuration with stiffness modulation demonstrated between two states: the atmospheric state and the jammed vacuum state.
The 6-in-1 circular pattern exhibited the highest dynamic range and the most uniform spatial distribution of stiffness. 
Consequently, it is an ideal candidate for the recreation of tactile sensations associated with organs that possess high mechanical variation.
Although the hexagonal bipyramid and octahedral patterns showed a higher spatial variance, they still offer a high baseline stiffness that could mimic deeper, more rigid internal pathologies.

These findings suggest that the integration of 3D-printed structured fabrics with vacuum jamming is a promising and practical approach for modulating tactile sensations of stiffness.

The ability to span the sub-megapascal to megapascal threshold enables the VSTaI to effectively simulate the mechanical characteristics of various soft tissues, while its shape conformability allows it to serve as a variable-stiffness interface when overlaid on a soft internal core, with the perceived stiffness resulting from the combined mechanical properties of all materials.

\section{Conclusion}

In this study, the VSTaI capable of dynamically rendering tunable stiffness with the capability of shape conformability for palpation training was designed, fabricated, and experimentally characterized.
The fabrics achieve high shape adaptability through their 3D-printed interlocking architecture, which allows the individual links to rotate and slide relative to one another before the vacuum is applied.
Stiffness was modulated through vacuum-induced jamming of the 3D-printed structured fabric layer that was sealed in a membrane and sandwiched between two compliant silicone layers. 
Four interlocking fabric patterns with two geometric parameter sets were characterized using force–indentation tests at five locations on the surface. 
Across all configurations, a clear stiffness increase was found under vacuum jammed condition, yielding effective moduli ranging from the upper sub-megapascal level to the megapascal level. 
The largest change was observed in the 6-in-1 circular configuration, where the mean stiffness rose from 506.62 kPa to 1220.05 kPa, demonstrating a 140\% increase. 
Spatial uniformity was also investigated.
The circular chainmail patterns showed the most consistent stiffness across the interface, while the octahedral and hexagonal bipyramid patterns exhibited higher variability, but also provided higher baseline stiffness that may be useful when representing more rigid internal features.

Overall, these results suggest that structured-fabric jamming can offer a practical approach for tunable, near-soft-tissue stiffness rendering in a tangible and shape adaptable form factor, with pattern selection over both dynamic range and spatial consistency. 
Although these results establish a baseline for indentation stiffness, further studies are needed to establish the relationship between air pressure versus rigidity, and to explore more comprehensive mechanical behaviors, such as viscoelasticity that influence the tactile experience.
In addition, the present study reports a stiffness range obtained across multiple fabricated pattern topologies and geometric parameter sets, with each sample evaluated only under two operating conditions: the atmospheric state and the jammed vacuum state.
Future work will include dynamic and repeated-loading tests to characterize the response under intermediate stiffness states, and to quantify local variability and trial-to-trial repeatability more rigorously. We will also aim to expand the design space to a wider range of structural fabric geometries and parameters, as well as conduct user studies with representative VSTaI configurations to assess stiffness discriminability during palpation and its relevance to training applications.

\begin{credits}
\subsubsection{\ackname} The work presented in this paper and Dr Yiting Mo are funded by the Duke-NUS Research Support Fund.

\end{credits}

\end{document}